\documentclass[review]{elsarticle}
\graphicspath{{./figures/}}

\usepackage{hyperref}
\usepackage{float}
\usepackage{verbatim} 
\usepackage{apalike}
\usepackage{algorithm}
\usepackage{algorithmicx}
\usepackage{tabularx}
\usepackage{amsmath}
\usepackage{adjustbox}
\usepackage{xcolor}%
\usepackage{multirow}
\usepackage{multicol}
\usepackage{csquotes}

\restylefloat{figure}
\restylefloat{table}

\journal{Expert Systems with Applications}

\biboptions{authoryear}

\begin{document}
\begin{frontmatter}

\begin{titlepage}
\begin{center}
\vspace*{1cm}

\textbf{ \large Strategic Prompting for Conversational Tasks: A Comparative Analysis of Large Language Models Across Diverse Conversational Tasks}

\vspace{1.5cm}

Ratnesh Kumar Joshi$^a$ (ratnesh\_1921cs28@iitp.ac.in), \\Priyanshu Priya$^a$ (priyanshu\_20212cs26@iitp.ac.in), \\Vishesh Desai$^a$ (desai\_2211ai07@iitp.ac.in), \\Saurav Dudhate $^a$ (saurav\_2001cs62@iitp.ac.in), \\Siddhant Senapati$^a$ (siddhant\_2101ai38@iitp.ac.in), \\Asif Ekbal$^a$ (asif@iitp.ac.in),  \\Roshni Ramnani$^b$ (roshni.r.ramnani@accenture.com), \\Anutosh Maitra$^b$ (anutosh.maitra@accenture.com), \\Shubhashis Sengupta$^b$ (shubhashis.sengupta@accenture.com) 

\hspace{10pt}

\begin{flushleft}
\small  
$^a$ Department of Computer Science and Engineering, IIT Patna, Patna, India \\
$^b$ Technology Innovation Lab, Accenture, Bengaluru, India \\

\vspace{1cm}
\textbf{Corresponding Author:} \\
Ratnesh Kumar Joshi \\
Department of Computer Science and Engineering, IIT Patna, Patna, India \\
Tel: +91 8979773589 \\
Email: ratnesh\_1921cs28@iitp.ac.in; ratneshkr.joshi@gmail.com

\end{flushleft}        
\end{center}
\end{titlepage}

\title{Strategic Prompting for Conversational Tasks: A Comparative Analysis of Large Language Models Across Diverse Conversational Tasks}

\author[label1]{Ratnesh Kr. Joshi \corref{cor1}}
\ead{ratnesh\_1921cs28@iitp.ac.in}

\author[label1]{Priyanshu Priya}
\ead{priyanshu\_20212cs26@iitp.ac.in}

\author[label1]{Vishesh Desai}
\ead{desai\_2211ai07@iitp.ac.in}

\author[label1]{Saurav Dudhate}
\ead{saurav\_2001cs62@iitp.ac.in}

\author[label1]{Siddhant Senapati}
\ead{siddhant\_2101ai38@iitp.ac.in}

\author[label1]{Asif Ekbal}
\ead{asif@iitp.ac.in}

\author[label2]{Roshni Ramnani}
\ead{roshni.r.ramnani@accenture.com}
\
\author[label2]{Anutosh Maitra}
\ead{anutosh.maitra@accenture.com}

\author[label2]{Shubhashis Sengupta}
\ead{shubhashis.sengupta@accenture.com}

\cortext[cor1]{Corresponding author.}
\address[label1]{Department of Computer Science and Engineering, IIT Patna, Patna, India}
\address[label2]{Technology Innovation Lab, Accenture, Bengaluru, India }

\begin{abstract}
Given the advancements in conversational artificial intelligence, the evaluation and assessment of Large Language Models (LLMs) play a crucial role in ensuring optimal performance across various conversational tasks. In this paper, we present a comprehensive study that thoroughly evaluates the capabilities and limitations of five prevalent LLMs: Llama, OPT, Falcon, Alpaca, and MPT. The study encompasses various conversational tasks, including reservation, empathetic response generation, mental health and legal counseling, persuasion, and negotiation. To conduct the evaluation, an extensive test setup is employed, utilizing multiple evaluation criteria that span from automatic to human evaluation. This includes using generic and task-specific metrics to gauge the LMs' performance accurately. From our evaluation, no single model emerges as universally optimal for all tasks. Instead, their performance varies significantly depending on the specific requirements of each task. While some models excel in certain tasks, they may demonstrate comparatively poorer performance in others. These findings emphasize the importance of considering task-specific requirements and characteristics when selecting the most suitable LM for conversational applications.

\end{abstract}
\begin{keyword}
LLMs \sep Evaluation \sep Conversational Agents
\end{keyword}
\end{frontmatter}

\section{Introduction}\label{Introduction}

Large Language Models (LLMs) (a.k.a generative AI) are language models with parameter sizes over a hundred billion. Examples of LLMs include Megatron \cite{shoeybi2019megatron}, GPT-3 \cite{brown2020language}, Gopher \cite{rae2021scaling}, GPT-Jurassic \cite{lieber2021jurassic}, OPT-175B \cite{zhang2022opt}, Falcon \cite{zxhang2023falcon}, Llama \cite{touvron2023Llama}, InstructGPT \cite{ouyang2022training}, ChatGPT\footnote{\href{https://chat.openai.com/}{https://chat.openai.com/}}, etc. In recent times, there has been a significant surge in the prominence of LLMs. With the release of ChatGPT, they have gained widespread recognition and use across various domains, including research and industry; perhaps most notably, they have become more accessible and well-known to the general public. These unsupervised autoregressive models can forecast the next tokens, whether characters, words, or strings, by leveraging contextual information from previous data. To enable the LLMs to demonstrate their abilities, sophisticated prompt engineering \cite{neuralmagic2023} is required. Prompts are used to probe the LLMs to generate the target outcome by sampling the language distribution. 

The rapid advancements and impressive achievements of LLMs have triggered a revolutionary shift in Natural Language Processing (NLP). These models have showcased substantial improvements across a variety of traditional tasks in NLP, including natural language understanding tasks like emotion recognition \cite{zhang2023refashioning,zhao2023chatgpt}, hate speech detection \cite{huang2023chatgpt,oliveira2023good}, to name a few, to generative tasks such as summarization \cite{liang2022holistic,pu2023chatgpt}, dialogue \cite{lin2023llm}, code generation \cite{liu2023your}, and more.

Large language models have many applications, including language generation for articles, essays, and creative writing \cite{swanson2021story}, as well as translation \cite{xie2023translating}, paraphrasing \cite{tang2023not}, and summarization \cite{shen2023large} tasks. They can power chatbots and virtual assistants capable of understanding and responding to natural language queries while also serving as question-answering systems \cite{robinson2022leveraging}. Text classification and sentiment analysis benefit from these models, aiding in tasks like customer review sentiment classification \cite{zhang2023sentiment} and spam detection. 
The applications of large language models are diverse and expand across various domains.

While large language models excel at many tasks, it is important to highlight that they may not always perform exceptionally well for every task. While they possess impressive language processing capabilities, they might struggle with certain specialized or domain-specific tasks that require specific knowledge or expertise \cite{omiye2023large, thapa2023chatgpt}. It is crucial to carefully evaluate the strengths and limitations of LLMs for each task and consider other techniques or approaches if needed. Additionally, fine-tuning the model on specific data or using task-specific architectures can help optimize performance for individual tasks.

To gain a deep understanding of the capabilities and performance of different large language models across various tasks, extensive evaluation becomes necessary. Evaluating these models on various tasks allows us to investigate their strengths and weaknesses in different domains. Through rigorous evaluation, we try to identify which large language models work best for specific tasks to enable researchers and developers to make informed decisions and choose the most suitable model based on the task requirements. The proposed evaluation process hopefully can help in advancing the field of language models by fostering a better understanding of their applicability and performance characteristics.

In this work, we explore the performance of 5 different LLMs,  Llama \cite{touvron2023Llama}, OPT \cite{zhang2022opt}, Falcon \cite{zxhang2023falcon}, Alpaca \cite{alpaca} and MPT \cite{MosaicML2023Introducing} across a diverse range of tasks, namely reservation, empathetic response generation, mental health and legal counseling, persuasion, and negotiation.

\section{Related Work}\label{RelatedWorks}

In recent years, Large Language Models (LLM) have emerged as a focal point of extensive research and development within natural language processing (NLP) \cite{zhao2023survey, chang2023survey}.  Google's T5 \cite{raffel2020exploring}, introduced in 2019, marked a pivotal moment with its text-to-text approach, departing from the reliance on task-specific architectures. This innovative approach offered unparalleled flexibility and adaptability across various NLP tasks. Shortly thereafter, OpenAI introduced GPT-3 \cite{brown2020language}, which not only showcased remarkable scale and performance but also set new benchmarks, propelling the field into uncharted territory and sparking tremendous interest and investment in LLM research.

Subsequent years have witnessed a remarkable boom in the development and release of numerous other LLMs, each contributing to the diversification and enhancement of NLP capabilities. Among these models, ChatGPT\footnote{\href{https://chat.openai.com/}{https://chat.openai.com/}}, a conversational model, has gained prominence for its ability to engage in human-like interactions and generate coherent dialogues. Additionally, Facebook's text model Llama \cite{touvron2023Llama} has made significant contributions to the NLP ecosystem, leading to comprehensive models such as Falcon \cite{zxhang2023falcon} and Alpaca \cite{alpaca}, offering robust language understanding capabilities and addressing complex language comprehension tasks. These models can be used in a wide variety of tasks such as machine translation \cite{GAO2024125087}, code simplification \cite{PATSAKIS2024124912}, medical diagnosis \cite{CARUCCIO2024121186}, and social media analysis \cite{NAJAFI2024124737}. These models continue to shape the NLP landscape, pushing the boundaries of research and application possibilities in natural language processing and opening new horizons for innovation and discovery.

The training of large-scale language models \cite{li2023flm, nijkamp2023codegen2} is a multifaceted process that unfolds in two distinct phases: pretraining \cite{erhan2010does} and fine-tuning \cite{peng2023instruction}. During the pretraining phase, these models build a foundational understanding of language, encompassing aspects such as grammar, factual knowledge, and some rudimentary reasoning capabilities. This is achieved through exposure to vast volumes of textual data, where the models learn to recognize patterns, linguistic structures, and the intricacies of human language.  Fine-tuning, the second phase of training represents tailoring of these models to specific tasks or applications. In this stage, the models' general language understanding is channeled towards particular objectives, such as text generation, question answering, chatbot interactions, or following specific instructions. Fine-tuning enables these models to specialize and adapt their acquired linguistic knowledge to excel in these specific tasks \cite{jang2023exploring, chen2020adversarial}, thus enhancing their practical utility and relevance in real-world scenarios. Furthermore, the incorporation of Reinforcement Learning from Human Feedback (RLHF) \cite{griffith2013policy} during fine-tuning iteratively refines the models' responses and behavior, ensuring they align more closely with desired outcomes and continue to improve their effectiveness in addressing practical language-related challenges.

Our work primarily focuses on the prompting-based evaluation of select LLMs based on a set of predefined properties that are important for a conversational agent. These properties include persuasion, negotiation, information retrieval, politeness, and empathy. We employ a total of five dataset ( detailed in section \ref{Datasets}) to assess these desired properties.

Current LLms have a few gaps which pose challenges to their practical usability and reliability. One such prominent limitation is the tendency of LLM-generated responses to be excessively verbose \cite{aiyappa2023can}, often containing unwanted preambles and post-amble phrases such as \enquote{I see that you want to...} or \enquote{So to summarize...}. This verbosity can compromise the clarity and conciseness of the generated text. To combat this issue, practitioners employ techniques like careful prompting and providing few-shot examples. In this work, by structuring input prompts or examples strategically, we guide LLMs to produce more concise and relevant responses, effectively curbing unnecessary verbosity and ensuring that generated content aligns better with the desired communication style. Another critical challenge arises from the LLM's occasional tendency to generate information that may not be accurate or factual, essentially \enquote{making things up}. \cite{rawte2023survey, LUO2024124456} This can lead to potentially misleading or erroneous responses in specific contexts. Addressing this issue requires rigorous fact-checking and post-processing of model outputs to filter out inaccuracies. To this end, we incorporate prompting LLMs with domain-specific data and explicit guidelines for generating accurate content, essential to enhance factual accuracy and reliability. By collectively tackling these known gaps through informed input strategies, fact-checking, and fine-tuning, LLMs can become more dependable tools in various real-world applications.

\section{Methodology and Experimental Setup}\label{Methodology}

\subsection{Tasks/Datasets}\label{Datasets}

In this section, we provide an overview of the datasets used in our evaluation of various Large Language Models (LLMs) through zero-shot, one-shot, and two-shot learning prompts. These datasets have been carefully selected to assess the capabilities of LLMs in different linguistic and conversational domains. Each dataset is a unique benchmark for evaluating the model's performance in a specific context. The 5 following datasets were chosen for our evaluation work: 

\textbf{Multi-Domain Wizard-of-Oz} or MultiWOZ \cite{zang2020MultiWOZ} is a multi-domain dialogue dataset. In total, the dataset consists of 7 domains - Attraction, Hospital, Police, Hotel, Restaurant, Taxi, and Train. The latter four are extended domains, which include the sub-task Booking. The dataset has 10438 dialogues, with 8438 dialogues for training and 1000 dialogues each for evaluation and testing.


The \textbf{Craigslist Bargain} \cite{he2018decoupling} dataset comprises textual content extracted from Craigslist listings, a platform where individuals advertise items for sale or trade. The dataset consists of 6682 conversations between people using Amazon Mechanical Turk (AMT). The dataset incorporates a variety of persuasive and negotiating strategies like embellishment, side offers, and appeals to sympathy. This dataset is widely utilized for various Natural Language Processing (NLP) tasks, including text classification, sentiment analysis, and recommendation systems. 


The \textbf{Mental Health and Legal Counselling Dataset (MHLCD)} \cite{mishra2023help} includes conversations about mental health and legal counseling support for women and children who have been the victims of any kind of crime, such as domestic violence, rape, acid attack, etc. The MHLCD dataset is annotated with three informative labels, viz. counseling strategies, politeness, and empathy. The dataset contains 755 dialogues for training, 100 dialogues for evaluation, and 151 dialogues for testing.

The \textbf{Empathetic Dialogues} \cite{rashkin2019towards} dataset is a large-scale multi-turn empathetic dialogue dataset collected on the Amazon Mechanical Turk, containing 24,850 one-to-one open-domain conversations. The dataset contains 25k conversations grounded in emotional situations. This dataset's methodology is based on three tasks: \textit{Empathy/Sympathy:} Did the responses show understanding of the feelings of the person talking about their experience? \textit{Relevance:} Did the responses seem appropriate to the conversation? Were they on-topic? \textit{Fluency:} Could you understand the responses? Did the language seem accurate? 


The \textbf{Empathetic Persuasions} \cite{samad2022empathetic} dataset is an empathy annotated version of Persuasion for Good dataset \cite{wang2019persuasion}. To annotate the Empathetic Persuasions dataset with different emotion labels, authors used the Empathetic Dialogues \cite{rashkin2019towards} dataset consisting of 25k dialogues grounded with 32 different emotions. The dataset contains 1017 conversations.


\subsection{Prompting}\label{Prompting}

We have created prompts that we provide to the LLMs to get the responses. The structure of the prompt is as follows:\\

\noindent \textbf{System Prompt:} Here, the system prompt is the generalized part of the prompt, and this has been used in all the prompts for all the tasks. It guides the Language Model to generate the responses in a Conversational Setting. The prompt structure is as follows, and the missing details are filled with instructions specific to the task.\\
\enquote{[SYSTEM] Your task is to generate coherent and contextually relevant responses based on the given input. Your responses should aim to ...
The goal is to... Please ensure the response is grounded to the last user input and context.}

\noindent\textbf{Few-Shot:} We have added few-shot examples for each context response pair in the test set. For each test sample in the test set, we sample 4 context-response pairs. All context-response pairs correspond to a similar type of Generated Response. For each test sample, 4 prompts have been created as follows: \textit{Zero-shot} = only context used to generate the response, \textit{One-shot} = 1 example along with the context used to generate the response, \textit{Two-shot} = 2 examples along with the context used to generate the response.

\noindent\textbf{Context:} The context is a sequence of user and system utterances corresponding to the dialogue. The last utterance in this sequence is the user utterance, and the Language Model predicts the next System utterance considering the few-shot examples and the context provided regarding dialogue history.

\noindent\textbf{Gold Response:} For each test sample, the next System utterance after the last utterance of the context, according to the dialogue in the data set, has been taken as the Gold Response. We have used this Gold Response to evaluate some of the generic automatic metrics (token-based similarity like BLEU, METEOR, and embedding-based similarity like BERT Score, Greedy Matching, etc.; more details are mentioned in Evaluation Setup).

Prompt specific to an Evaluation Metric: In this paper, apart from the general format of prompts mentioned above, a slightly different format of prompt has also been used to evaluate Context Consistency for MultiWOZ dialogue dataset.

For each test sample, the entire Context remains the same, but the last User utterance in the context is changed. Except for the User query or the last user utterance, everything remains constant across all the rested models, including the system prompt and the context. This way, for each dialogue (test sample), we create 3 prompts for 3 alternate user queries. 

\subsection{Large Language Models}\label{Models}

In this research study, we delve into the evaluation of five state-of-the-art Large Language Models (LLMs) within the context of generating text in response to zero-shot, one-shot, and two-shot prompts. The evaluated LLMs are Falcon, OPT, MPT, Llama, and Alpaca. We aim to assess their capabilities in understanding, generalizing, and responding to various prompts in diverse datasets, shedding light on their performance across various tasks and domains.

Falcon \cite{zxhang2023falcon} Falcon LLM has been trained on 1 trillion tokens and is available in various versions having 180B, 40B, 7.5B, and 1.3B parameters, respectively. We utilize the Falcon 7B Instruct version\footnote{Model available at \url{https://huggingface.co/tiiuae/falcon-7b-instruct}.} for our experiments.
 
The Open Pre-trained Transformer Language Models \cite{zhang2022opt} or OPT model was proposed in Open Pre-trained Transformer Language Models by Meta AI. OPT is a series of open-sourced large causal language models that perform similarly to GPT-3. Currently, six models are available on Hugging Face: 125M, 350M, 1.3B, 2.7B, 6.7B, and 30B parameters. OPT was trained on publicly available data sets to allow for more community engagement in understanding this foundational new technology. We utilize the OPT 6.7B Instruct version\footnote{Model available at \url{https://huggingface.co/facebook/opt-6.7bt}.} for our experiments.

The MosaicML Pretrained Transformer \cite{MosaicML2023Introducing} or MPT models are GPT-style decoder-only transformers that come with many improvements, such as performance-optimized layer implementations, greater training stability due to architecture changes and no context length limitations. MPT-7B is a transformer model trained from scratch using 1T tokens of text and code. It was trained on the MosaicML platform, with a time frame of 9.5 days with zero human intervention. MPT-7B is the base model and has been shown to outperform other open-source 7B - 20B models. We utilize the MPT-7B Instruct version\footnote{Model available at \url{https://huggingface.co/mosaicml/mpt-7b}.} for our experiments.

Llama \cite{touvron2023Llama} stands for Large Language Model Meta AI. It is a state-of-the-art foundational large language model by Meta AI designed to help researchers advance their work in this subfield of AI. Llama 2 pretrained models are trained on 2 trillion tokens and have to double the context length of Llama 1. Its fine-tuned models have been trained on over 1 million human annotations. For the current evaluation tasks, we have used the Llama 2 model. As a foundation model, Llama is designed to be versatile and can be applied to many different use cases versus a fine-tuned model that is designed for a specific task. We utilize the Llama2-7B version\footnote{Model available at \url{https://huggingface.co/meta-Llama/Llama-2-7b-hf}.} for our experiments.

Alpaca \cite{alpaca} is an instruction-following language model, which is fine-tuned from Meta’s Llama 7B model. It is trained on 52K instruction-following demonstrations generated in the style of self-instruct using text-davinci-003. On the self-instruct evaluation set, Alpaca shows many behaviors similar to OpenAI’s text-davinci-003 but is also surprisingly small and easy/cheap to reproduce.

\section{Evaluation Setup}\label{EvaluationSetup}

For the comprehensive evaluation of LLMs, we have used a set of Generic Metrics, which evaluates the general aspects of any conversational agent (like fluency context relevance). We have also used Task-specific Metrics, which evaluate the ability of LLMs to perform a particular task (like negotiation, empathy, etc.).

In this paper, we present the two approaches for evaluation: 1. Automatic Evaluation and 2. Human Evaluation. 

\subsection{Automatic Evaluation}\label{AutomaticEvaluation}

The metrics presented in this paper to perform automatic evaluation are divided into two segments: 1. Generic Metrics  2. Task-specific Metrics

\subsubsection{Generic Metrics}\label{AutomaticGenericMetrics}

The generic metrics for automatic evaluation presented in this paper can be further divided into following buckets:\\

    \noindent \textbf{1. Lexical-based similarity} between generated responses and gold responses: \textit{BLEU} \cite{papineni2002bleu} (BLEU1, BLEU2, BLEU3, BLEU4), \textit{METEOR} \cite{banerjee2005meteor}\\ 

    \noindent \textbf{2. Embedding-based similarity} between generated responses and gold responses: \textit{Greedy Matching} \cite{rus2012optimal}, \textit{Embedding Average} \cite{landauer1997solution}, \textit{Vector Extrema} \cite{forgues2014bootstrapping}, \textit{BertScore} \cite{zhang2019bertscore}\\



\subsubsection{Task-specific Metrics}\label{AutomaticTaskMetrics}

    \noindent \textbf{1. MultiWOZ Dialogue Dataset}: This dataset evaluates the ability of a LLMs to capture the requested information mentioned in the user query and provide relevant and useful information in the generated response. In \textbf{Context Consistency}, we compute the similarity among the three alternate queries as follows: $(similarity(q_1,q_2)+similarity(q_2,q_3)+similarity(q_3,q_1))/3$, where $q_1$,$q_2$, and $q_3$ are the generated responses for the queries, and similarity($q_i$,$q_j$) computes the cosine similarity between the Bert \cite{devlin2018bert} embeddings for the responses $q_i$ and $q_j$. \textbf{INFORM} evaluates whether the slot values informed by the user appear in the system-generated response or not. More weights are given to newly added slot values compared to existing slot values. \textbf{SUCCESS} evaluates whether the slot values requested by the user appear in the system-generated response. This gives us the SUCCESS metric score; the higher the score, the higher the deviation, and worse the performance.

    \noindent \textbf{2. CraiglistBargain}: Here, we have used the mentioned task-specific metrics to evaluate the ability of the Language Model to negotiate with the user based on price. \textbf{NegStr} computes the percentage of negotiating utterances. We train a BERT \cite{devlin2018bert} based binary classifier model based on the annotation present in the CraiglistBargain dataset. \textbf{BarStr:} computes how close the generated price is to the seller's original price while remaining within the limit of the current buyer and seller prices as lower and upper bounds, respectively. \\

    \noindent \textbf{3. MHLCD Dataset.} The effectiveness of LLMs in terms of task success on the MHLCD dataset is evaluated using three task-specific metrics, \textit{viz.} CoStr, Pol and Emp. \textit{Costr} indicates the number of utterances generated by the LLMs with counseling strategy. \textit{Pol} and \textit{Emp} represent the number of polite and empathetic utterances the LLMs generate. \textbf{CoStr} computes the percentage of utterances generated with a counseling strategy (checks whether counseling strategy is present or absent). \textbf{Pol} computes the percentage of polite utterances generated. \textbf{Emp} computes the percentage of empathetic utterances generated.\\

    \noindent \textbf{4. Empathetic Persuasions:} The metrics evaluate how well a Language Model can persuade the user by generating emotional and empathetic responses. \textbf{PerStr} computes the percentage of persuasive utterances. \textbf{EmoPr} computes the percentage of empathetic utterances generated.\\

    \noindent \textbf{5. Empathetic Dialogues:} For this dataset, our focus is on the ability of the Language Model to showcase empathy towards the user. To this end, the metric \textbf{Emp} computes the percentage of empathetic utterances generated out of the total generated responses.\\

\subsection{Human Evaluation} \label{HumanEvaluation}
To assess the responses generated by the LLMs regarding response quality and task success, we also conduct a human evaluation of the generated responses for all the datasets. 

\subsubsection{Generic Metrics}\label{HumanGenericMetrics}
Given below are the evaluation metrics to evaluate the general aspects of a conversational agent:

    \noindent \textbf{1. Fluency}: It refers to the correctness of the generated response concerning grammar and word choice, including spelling. The score in the range of 1 to 5 has been given to each test sample (generated response). The scoring is 1 (Poor), 2 (Fair), 3 (Average),4 (Good), and 5 (Excellent).

    \noindent \textbf{2. Context Relevance}: The response should correctly address the query and be consistent with the contents of the context provided. If the response follows the flow of the given context, the response should be logically sound for the user query. Here also, we have given scores in the range of 1 to 5. The scoring criteria are: 1 - response completely inconsistent with the context and provide no useful information, 2 - response provides some random information that is not at all asked for in the context, 3 - response provides only some of the information asked for in the context with some extra irrelevant information, 4 - response provides only some of the information asked for in the context without adding any extra irrelevant information, 5 - response provides all the relevant information asked for in the context without adding any extra irrelevant information.

    \noindent \textbf{3. Non-Repetitiveness}: to check the non-repetitiveness of the generated utterance. The metrics score higher in case the response does not include repeated words.

\subsubsection{Task-specific Metrics}\label{HumanTaskMetrics}

Here, for each dataset, we present dedicated human evaluation metrics to evaluate the Language Model for the tasks associated with these datasets.

    \noindent \textbf{1. MHLCD. }For human evaluation, we employ three task-specific metrics to evaluate the performance of the LLMs, namely Counseling strategy correctness (Con), Politeness (Pol), and Empathy (Emp). The human evaluators rate the responses generated by the LLMs on an integer Likert scale of 1-5\footnote{The scale 1-5 denotes low to high intensity, such as Con = 1 denotes non-counseling and Con = 5 denotes highly counseling.}.

    \noindent\textbf{2. Empathetic Persuasion:} In the human evaluation of the Empathetic Persuasions dataset, we utilize three key metrics, each of which serves as an essential dimension in understanding the persuasive quality of the generated dialogue. \textbf{Persuasiveness (Per)} assesses the overall persuasiveness of the generated dialogue. Human evaluators rate the dialogue on a one-to-five positive integer scale, where a higher score indicates a more persuasive response. \textbf{Empathy (Emp)} plays a crucial role in persuasive communication. Evaluators rate the dialogue based on the degree of empathy conveyed. 
    \textbf{Donation Probability (DonPr)} This metric quantifies the real-world impact of persuasive dialogue. By calculating the percentage of time people donated after engaging with the generated content, we can measure the effectiveness of LLMs in motivating actions and achieving the desired outcome. 

    \noindent \textbf{3. Empathetic Dialogues:}  The primary motivation for conducting a human evaluation on this dataset is to assess the effectiveness of LLMs in empathetic/sympathetic communication.  In the human evaluation of the Empathetic Dialogues dataset, \textbf{Empathy/Sympathy:} plays a crucial role in persuasive communication. Evaluators rate the dialogue based on the degree of empathy conveyed. Like persuasiveness, this metric also employs a one-to-five positive integer scale.

    \noindent \textbf{4. CraiglistBargain:} For this dataset, the evaluation primarily focused on negotiation capability and some degree of logical understanding of the product when suggesting the price. \textbf{Negotiation Consistency} corresponds with how well the negotiation strategy (bargaining intensity) relates to the context. \textbf{Dialogue Engagement} corresponds with how well the response makes sense as a response and how well it promotes conversation between the two parties.

\subsubsection{Metrics Details}\label{MetricDetails}

    \textbf{1. MultiWOZ Dialogue Dataset}: This dataset evaluates the ability of a Language Model to capture the requested information mentioned in the user query and provide relevant and useful information in the generated response.
    Given below are the metrics used for the evaluation of the MultiWOZ dataset:\\

    \textbf{a. Context Consistency}: User queries in goal-oriented dialogues are less predictable. Different users can request the same information (mentioned in the context) and can ask to perform the same task but can ask in different ways. To simulate this human behavior, for each test sample, we create three alternate user queries providing the same context. Example:\\
    Context: USER: I am looking for a train to Boston from New York that leaves at 9 in the morning.\\
    SYSTEM: Two trains are leaving from New York for Boston, suitable for your timings, first at 9:15 a.m. and second at 9:40 a.m. Which train would you like to book ?\\
    USER1: Book one ticket for the 9:15 train\\
    USER2: Finalize my reservation for the one departing at 9:15\\
    USER3: Yes, the 9:15 one\\
    For each test sample, we compute the similarity among the three alternate queries as follows: $(similarity(q1,q2)+similarity(q2,q3)+similarity(q3,q1))/3$.
    Here, q1, q2, and q3 are the generated responses for the queries, and similarity(qi, qj) computes the cosine similarity between the Bert embeddings for the responses qi and qj. We compute the average of these similarity scores over all the test samples, and this is the Context Consistency score. The closer this score to 1, the higher the similarity among the alternate query responses and the more consistent the Language Model performance concerning the context.\\

    \textbf{b. INFORM}: In the MultiWOZ dataset, for each dialogue, there are annotations for the informed slots. These informed slots contain the information the user provides, mentioning the user's preferences and conditions as their values. The dialogue is considered to turn by turn till the intent of the user turn remains similar. For the dialogue, a set of newly added slots in the current utterance and a set of slots provided till the current utterance will be maintained.INFORM evaluates whether the slot values informed by the user appear in the system-generated response or not. More weights are given to newly added slot values compared to existing slot values. For a particular dialogue, the average matching score for each system utterance will be reported. For a Language Model, the aggregate matching score over all the dialogues is considered for comparison.\\
    \textbf{c. SUCCESS}: Similar to informed slots, the dataset also contains the annotations for requested slots. These slots consist of the information that the user is requesting from the system. Here, we consider the information requested to be a named entity. To roughly evaluate how well a Language Model provides the requested information, we compare the number of named entities in the generated response to the number of named entities in the gold response. For comparison, we adopt the following approach:\\
    For each test sample, we compute the number of named entities in the generated response (Nhi) and in gold/reference response (Nri). Then, we compute the absolute difference between Nhi and Nri ( di = abs(Nhi-Nri) ). We also compute the average number of named entities in the generated response (Nh = mean(Nhi)) and the Gold response (Nr = mean(Nri)) over the entire test set. Thus, we have a mean deviation in the number of entities ( mean(di) ) and an absolute deviation in the mean number of entities in generated and gold response (abs(Nh-Nr) ). Finally, we compute the average of mean deviation and absolute deviation of mean - (average(di) + absolute(average(Nri) - average(Nhi)))/2. This gives us the SUCCESS metric score; the higher the score, the higher the deviation, and worse the performance.

    \textbf{2. CraiglistBargain}: Here, we have used the mentioned task-specific metrics to evaluate the ability of the Language Model to negotiate with the user based on price:\\
    \textbf{a. NegStr:} It computes the percentage of negotiating utterances. We train a BERT \cite{devlin2018bert} based binary classifier model based on the annotation present in the CraigslistBargain dataset. \\
    \textbf{b. BarStr:} It computes how close the generated price is to the seller's original price while remaining within the limit of the current buyer and seller prices as lower and upper bounds, respectively. \\

    \textbf{3. MHLCD Dataset.} The effectiveness of LLMs in terms of task success on the MHLCD dataset is evaluated using three task-specific metrics, \textit{viz.} CoStr, Pol and Emp. \textit{Costr} indicates the number of utterances generated by the LLMs with counseling strategy. \textit{Pol} and \textit{Emp} represent the number of polite and empathetic utterances the LLMs generate.

    \textbf{4. Empathetic Persuasions:} The metrics evaluate how well a Language Model can persuade the user by generating emotional and empathetic responses.\\
    \textbf{a. PerStr:} Computes the percentage of persuasive utterances\\
    \textbf{b. EmoPr:} Computes the percentage of empathetic utterances generated.\\

    \textbf{5. Empathetic Dialogues:} Evaluates the ability of the Language Model to show empathy towards the user.\\
    \textbf{a. Emp:} It computes the percentage of empathetic utterances generated.\\

\section{Results and Analysis}\label{ResultsAndAnalysis}

\subsection{Results}\label{Results}

For our evaluation, we give much higher importance to the task-specific metrics and the human evaluation metrics over the Generic Automatic evaluation metrics \ref{ResultsGenericAutomatic}. This is primarily due to our focus being on the task performance of the models and also the lack of certainty of the gold response as the only valid response. There could be multiple valid responses for a context, so we give importance to the task-based and human evaluation metrics during evaluation. The task-specific metrics and the human evaluation metrics results are shown in Table \ref{human_results_mhlcd}, Table \ref{results_cb}, Table \ref{human_results_p4g}, Table \ref{human_results_empathetic_dialogues} and Table \ref{human_results_woz}. Based on the Evaluation results, no model performs best for all the tasks. Some models perform better on certain types of tasks.

\subsection{Results: Generic Automatic Evaluation}\label{ResultsGenericAutomatic}

The results for generic automatic evaluation of the LLMs are shown in Table\ref{gen_auto_mhlcd}, Table\ref{gen_auto_ep}, Table\ref{gen_auto_cb}, Table\ref{gen_auto_ed} and Table\ref{gen_auto_woz}

\begin{table*}[ht]
\centering
\begin{adjustbox}{max width=\linewidth}
\begin{tabular}{|l|l|l|l|l|l|l|l|l|l|l|l|}
\hline
\textbf{}       &           & \textbf{B1} & \textbf{B2} & \textbf{B3} & \textbf{B4} & \textbf{METEOR} & \textbf{ROUGE-L} & \textbf{Bert Score} & \textbf{Embedding Average} & \textbf{Vector Extrema} & \textbf{Greedy Matching} \\ \hline
\textbf{Llama2} & Zero-shot & 0.083       & 0.041       & 0.026       & 0.017       & 0.068           & 0.087            & 0.823               & 0.624                      & 0.273                   & 0.761                    \\ \hline
\textbf{}       & One-shot  & 0.082       & 0.04        & 0.025       & 0.016       & 0.069           & 0.088            & 0.824               & 0.629                      & 0.279                   & 0.763                    \\ \hline
\textbf{}       & Two-shot  & 0.088       & 0.042       & 0.026       & 0.018       & 0.07            & 0.09             & 0.832               & 0.639                      & 0.287                   & 0.767                    \\ \hline
\textbf{Falcon} & Zero-shot & 0.06        & 0.016       & 0.007       & 0.003       & 0.048           & 0.052            & 0.808               & 0.641                      & 0.332                   & 0.758                    \\ \hline
\textbf{}       & One-shot  & 0.073       & 0.024       & 0.01        & 0.005       & 0.051           & 0.06             & 0.811               & 0.666                      & 0.346                   & 0.769                    \\ \hline
\textbf{}       & Two-shot  & 0.068       & 0.019       & 0.007       & 0.003       & 0.038           & 0.048            & 0.799               & 0.653                      & 0.339                   & 0.756                    \\ \hline
\textbf{OPT}    & Zero-shot & 0.028       & 0.023       & 0.018       & 0.013       & 0.037           & 0.001            & 0.132               & 0.74                       & 0.419                   & 0.643                    \\ \hline
\textbf{}       & One-shot  & 0.029       & 0.021       & 0.016       & 0.009       & 0.041           & 0.001            & 0.495               & 0.692                      & 0.372                   & 0.74                     \\ \hline
\textbf{}       & Two-shot  & 0.018       & 0.014       & 0.013       & 0.007       & 0.036           & 0.001            & 0.025               & 0.746                      & 0.426                   & 0.626                    \\ \hline
\textbf{ALPACA} & Zero-shot & 0.171       & 0.091       & 0.058       & 0.038       & 0.106           & 0.143            & 0.859               & 0.869                      & 0.439                   & 0.749                    \\ \hline
\textbf{}       & One-shot  & 0.159       & 0.084       & 0.053       & 0.036       & 0.091           & 0.131            & 0.855               & 0.833                      & 0.42                    & 0.737                    \\ \hline
\textbf{}       & Two-shot  & 0.136       & 0.059       & 0.036       & 0.024       & 0.073           & 0.11             & 0.843               & 0.835                      & 0.404                   & 0.731                    \\ \hline
\textbf{MPT}    & Zero-shot & 0.183       & 0.105       & 0.066       & 0.042       & 0.099           & 0.16             & 0.864               & 0.875                      & 0.455                   & 0.753                    \\ \hline
\textbf{}       & One-shot  & 0.172       & 0.098       & 0.063       & 0.042       & 0.096           & 0.141            & 0.861               & 0.869                      & 0.445                   & 0.748                    \\ \hline
\textbf{}       & Two-shot  & 0.171       & 0.094       & 0.06        & 0.041       & 0.094           & 0.139            & 0.858               & 0.87                       & 0.44                    & 0.746                    \\ \hline
\end{tabular}
\end{adjustbox}
\caption{Results for generic automatic evaluation of \textbf{MHLCD} Dataset}
\label{gen_auto_mhlcd}
\end{table*}

\begin{table*}[ht]
\centering
\begin{adjustbox}{max width=\linewidth}
\begin{tabular}{|l|l|l|l|l|l|l|l|l|l|l|l|}
\hline
\textbf{}       &           & \textbf{B1} & \textbf{B2} & \textbf{B3} & \textbf{B4} & \textbf{METEOR} & \textbf{ROUGE-L} & \textbf{Bert Score} & \textbf{Embedding Average} & \textbf{Vector Extrema} & \textbf{Greedy Matching} \\ \hline
\textbf{Llama2} & Zero-shot & 0.093       & 0.044       & 0.022       & 0.01        & 0.075           & 0.131            & 0.152               & 0.849                      & 0.433                   & 0.731                    \\ \hline
\textbf{}       & One-shot  & 0.086       & 0.046       & 0.034       & 0.028       & 0.074           & 0.126            & 0.309               & 0.726                      & 0.385                   & 0.703                    \\ \hline
\textbf{}       & Two-shot  & 0.091       & 0.054       & 0.039       & 0.032       & 0.074           & 0.137            & 0.271               & 0.739                      & 0.387                   & 0.704                    \\ \hline
\textbf{Falcon} & Zero-shot & 0.067       & 0.033       & 0.019       & 0.01        & 0.067           & 0.122            & 0.152               & 0.834                      & 0.432                   & 0.728                    \\ \hline
\textbf{}       & One-shot  & 0.089       & 0.042       & 0.025       & 0.016       & 0.068           & 0.117            & 0.194               & 0.852                      & 0.423                   & 0.724                    \\ \hline
\textbf{}       & Two-shot  & 0.104       & 0.061       & 0.043       & 0.032       & 0.083           & 0.155            & 0.306               & 0.856                      & 0.45                    & 0.742                    \\ \hline
\textbf{OPT}    & Zero-shot & 0.12        & 0.06        & 0.035       & 0.02        & 0.076           & 0.147            & 0.177               & 0.826                      & 0.429                   & 0.732                    \\ \hline
\textbf{}       & One-shot  & 0.147       & \textbf{0.095}       & \textbf{0.075}       & \textbf{0.063}      & \textbf{0.098}           & 0.181            & 0.45                & 0.833                      & 0.475                   & 0.745                    \\ \hline
\textbf{}       & Two-shot  & \textbf{0.158}       & 0.093       & 0.068       & 0.054       & \textbf{0.098}           & \textbf{0.183}            & 0.464               & 0.835                      & 0.443                   & 0.74                     \\ \hline
\textbf{ALPACA} & Zero-shot & 0.102       & 0.047       & 0.024       & 0.01        & 0.074           & 0.106            & 0.112               & 0.67                       & 0.324                   & 0.679                    \\ \hline
\textbf{}       & One-shot  & 0.104       & 0.059       & 0.041       & 0.031       & 0.073           & 0.124            & 0.247               & 0.615                      & 0.31                    & 0.664                    \\ \hline
\textbf{}       & Two-shot  & 0.117       & 0.067       & 0.047       & 0.036       & 0.082           & 0.134            & 0.276               & 0.776                      & 0.387                   & 0.709                    \\ \hline
\textbf{MPT}    & Zero-shot & 0.122       & 0.061       & 0.033       & 0.019       & 0.083           & 0.156            & 0.273               & 0.869                      & 0.452                   & 0.741                    \\ \hline
\textbf{}       & One-shot  & 0.144       & 0.088       & 0.064       & 0.051       & 0.094           & 0.175            & \textbf{0.485}               & \textbf{0.884}                      & \textbf{0.477}                   & \textbf{0.758}                    \\ \hline
\textbf{}       & Two-shot  & 0.144       & 0.086       & 0.064       & 0.052       & 0.093           & 0.165            & 0.419               & 0.868                      & 0.454                   & 0.744                    \\ \hline
\end{tabular}
\end{adjustbox}
\caption{Results for generic automatic evaluation of \textbf{Empathetic Persuasions} Dataset}
\label{gen_auto_ep}
\end{table*}

\begin{table*}[ht]
\begin{adjustbox}{max width=\linewidth}
\centering
\begin{tabular}{|l|l|l|l|l|l|l|l|l|l|l|l|}
\hline
\textbf{}       &           & \textbf{B1} & \textbf{B2} & \textbf{B3} & \textbf{B4} & \textbf{METEOR} & \textbf{ROUGE-L} & \textbf{Bert Score} & \textbf{Embedding Average} & \textbf{Vector Extrema} & \textbf{Greedy Matching} \\ \hline
\textbf{Llama2} & Zero-shot & 0.09        & 0.037       & 0.019       & 0.01        & 0.075           & 0.083            & 0.856               & 0.778                      & 0.469                   & 0.682                    \\ \hline
\textbf{}       & One-shot  & 0.245       & 0.189       & 0.158       & 0.141       & 0.148           & 0.23             & 0.886               & 0.754                      & 0.486                   & 0.699                    \\ \hline
\textbf{}       & Two-shot  & 0.221       & 0.156       & 0.124       & 0.107       & 0.135           & 0.208            & 0.882               & 0.787                      & 0.502                   & 0.702                    \\ \hline
\textbf{OPT} & Zero-shot & 0.106       & 0.055       & 0.033       & 0.018       & 0.081           & 0.097            & 0.865               & 0.775                      & 0.49                    & 0.684                    \\ \hline
\textbf{}       & One-shot  & 0.231       & 0.165       & 0.133       & 0.115       & 0.136           & 0.232            & 0.883               & 0.712                      & 0.463                   & 0.683                    \\ \hline
\textbf{}       & Two-shot  & 0.187       & 0.122       & 0.096       & 0.083       & 0.11            & 0.157            & 0.87                & 0.668                      & 0.409                   & 0.65                     \\ \hline
\textbf{falcon}    & Zero-shot & 0.123       & 0.058       & 0.036       & 0.019       & 0.086           & 0.101            & 0.867               & 0.797                      & 0.51                    & 0.696                    \\ \hline
\textbf{}       & One-shot  & 0.277       & 0.22        & 0.188       & 0.168       & 0.174           & 0.282            & 0.897               & 0.862                      & 0.595                   & 0.758                    \\ \hline
\textbf{}       & Two-shot  & 0.227       & 0.165       & 0.137       & 0.121       & 0.153           & 0.227            & 0.889               & 0.832                      & 0.553                   & 0.732                    \\ \hline
\textbf{ALPACA} & Zero-shot & 0.096       & 0.042       & 0.023       & 0.013       & 0.077           & 0.082            & 0.858               & 0.794                      & 0.476                   & 0.675                    \\ \hline
\textbf{}       & One-shot  & 0.086       & 0.033       & 0.013       & 0.007       & 0.077           & 0.078            & 0.858               & 0.766                      & 0.449                   & 0.671                    \\ \hline
\textbf{}       & Two-shot  & 0.126       & 0.075       & 0.057       & 0.048       & 0.096           & 0.119            & 0.865               & 0.782                      & 0.48                    & 0.69                     \\ \hline
\textbf{MPT}    & Zero-shot & 0.132       & 0.06        & 0.034       & 0.018       & 0.086           & 0.107            & 0.865               & 0.813                      & 0.503                   & 0.698                    \\ \hline
\textbf{}       & One-shot  & 0.237       & 0.183       & 0.155       & 0.139       & 0.154           & 0.235            & 0.887               & 0.761                      & 0.495                   & 0.707                    \\ \hline
\textbf{}       & Two-shot  & 0.212       & 0.148       & 0.117       & 0.101       & 0.137           & 0.204            & 0.881               & 0.756                      & 0.485                   & 0.703                    \\ \hline
\end{tabular}
\end{adjustbox}
\caption{Results for generic automatic evaluation of \textbf{Craigslist Bargain} Dataset}
\label{gen_auto_cb}
\end{table*}

\begin{table*}[ht]
\centering
\begin{adjustbox}{max width=\linewidth}
\begin{tabular}{|l|l|l|l|l|l|l|l|l|l|l|l|}
\hline
\textbf{}       &           & \textbf{B1} & \textbf{B2} & \textbf{B3} & \textbf{B4} & \textbf{METEOR} & \textbf{ROUGE-L} & \textbf{Bert Score} & \textbf{Embedding Average} & \textbf{Vector Extrema} & \textbf{Greedy Matching} \\ \hline
\textbf{Llama2} & Zero-shot & 0.105       & 0.044       & 0.023       & 0.014       & 0.073           & 0.083            & 0.858               & 0.821                      & 0.474                   & 0.707                    \\ \hline
\textbf{}       & One-shot  & 0.099       & 0.039       & 0.021       & 0.013       & 0.077           & 0.085            & 0.862               & 0.828                      & 0.469                   & 0.706                    \\ \hline
\textbf{}       & Two-shot  & 0.078       & 0.014       & 0           & 0           & 0.067           & 0.068            & 0.859               & 0.816                      & 0.456                   & 0.691                    \\ \hline
\textbf{Falcon} & Zero-shot & 0.112       & 0.045       & 0.023      & 0.014       & 0.087           & 0.101           & 0.861               & 0.840                      & 0.486                    & 0.721                  \\ \hline
\textbf{}       & One-shot  & 0.106       & 0.043       & 0.023       & 0.012       & 0.080           & 0.088            & 0.862                & 0.845                      &0.482                   & 0.712                    \\ \hline
\textbf{}       & Two-shot  & 0.074       & 0.019        & 0.007       & 0           & 0.067           & 0.067            & 0.859               & 0.821                      & 0.440                    & 0.690                   \\ \hline
\textbf{OPT}    & Zero-shot & 0.096       & 0.044       & 0.025       & 0.016       & 0.065           & 0.084            & 0.856               & 0.795                      & 0.450                   & 0.701                    \\ \hline
\textbf{}       & One-shot  & 0.098       & 0.050       & 0.029       & 0.018       & 0.073           & 0.083            & 0.860                & 0.814                      & 0.465                   & 0.696                    \\ \hline
\textbf{}       & Two-shot  & 0.089       & 0.030       & 0.013      & 0       & 0.065           & 0.077            & 0.858               & 0.832                      & 0.460                   & 0.692                     \\ \hline
\textbf{ALPACA} & Zero-shot & 0.089       & 0.035       & 0.019       & 0.013       & 0.069           & 0.081            & 0.859               & 0.822                      & 0.449                   &  0.688                    \\ \hline
\textbf{}       & One-shot  & 0.089       &  0.027      & 0       & 0       & 0.066           & 0.074            & 0.855               & 0.826                      & 0.432                    &  0.681                    \\ \hline
\textbf{}       & Two-shot  & 0.089       & 0.031       & 0.016       & 0.007       & 0.071           & 0.074            & 0.856               & 0.767                      & 0.402                   & 0.667                    \\ \hline
\textbf{MPT}    & Zero-shot & 0.085       & 0.032       & 0.012       & 0.006       & 0.067           & 0.080            & 0.858               & 0.812                      & 0.492                   & 0.706                    \\ \hline
\textbf{}       & One-shot  & 0.081       & 0.028       & 0.013       & 0.007       & 0.065           & 0.070            & 0.859              & 0.837                      & 0.478                   & 0.708                   \\ \hline
\textbf{}       & Two-shot  & 0.095       & 0.030       & 0.012       & 0.006       & 0.071           & 0.080            & 0.860               & 0.838                     & 0.459                   & 0.696                   \\ \hline
\end{tabular}
\end{adjustbox}
\caption{Results for generic automatic evaluation of \textbf{Empathetic Dialogues} Dataset}
\label{gen_auto_ed}
\end{table*}

\begin{table*}[ht]
\centering
\begin{adjustbox}{max width=\linewidth}
\begin{tabular}{|l|l|l|l|l|l|l|l|l|l|l|l|}
\hline
\textbf{}       &           & \textbf{B1} & \textbf{B2} & \textbf{B3} & \textbf{B4} & \textbf{METEOR} & \textbf{ROUGE-L} & \textbf{Bert Score} & \textbf{Embedding Average} & \textbf{Vector Extrema} & \textbf{Greedy Matching} \\ \hline
\textbf{Llama2} & Zero-shot & 0.085       & 0.041       & 0.017       & 0.03        & 0.078           & 0.086            & 0.878               & 0.792                      & 0.495                   & 0.71                     \\ \hline
\textbf{}       & One-shot  & 0.263       & 0.202       & 0.173       & 0.187       & 0.196           & 0.32             & 0.912               & 0.813                      & 0.521                   & 0.738                    \\ \hline
\textbf{}       & Two-shot  & 0.256       & 0.187       & 0.154       & 0.128       & 0.167           & 0.273            & 0.903               & 0.824                      & 0.582                   & 0.767                    \\ \hline
\textbf{Falcon} & Zero-shot & 0.154       & 0.067       & 0.042       & 0.026       & 0.081           & 0.104            & 0.914               & 0.823                      & 0.58                    & 0.713                    \\ \hline
\textbf{}       & One-shot  & 0.278       & 0.182       & 0.145       & 0.128       & 0.152           & 0.286            & 0.922               & 0.789                      & 0.512                   & 0.724                    \\ \hline
\textbf{}       & Two-shot  & 0.19        & 0.132       & 0.104       & 0.092       & 0.18            & 0.163            & 0.95                & 0.698                      & 0.473                   & 0.72                     \\ \hline
\textbf{OPT}    & Zero-shot & 0.146       & 0.064       & 0.048       & 0.023       & 0.097           & 0.131            & 0.911               & 0.864                      & 0.58                    & 0.724                    \\ \hline
\textbf{}       & One-shot  & 0.296       & 0.34        & 0.192       & 0.177       & 0.186           & 0.295            & 0.933               & 0.912                      & 0.603                   & 0.772                    \\ \hline
\textbf{}       & Two-shot  & 0.234       & 0.178       & 0.142       & 0.129       & 0.162           & 0.235            & 0.922               & 0.864                      & 0.578                   & 0.768                    \\ \hline
\textbf{ALPACA} & Zero-shot & 0.102       & 0.053       & 0.032       & 0.019       & 0.083           & 0.088            & 0.892               & 0.812                      & 0.521                   & 0.699                    \\ \hline
\textbf{}       & One-shot  & 0.089       & 0.037       & 0.018       & 0.011       & 0.082           & 0.083            & 0.863               & 0.794                      & 0.457                   & 0.678                    \\ \hline
\textbf{}       & Two-shot  & 0.132       & 0.078       & 0.062       & 0.052       & 0.102           & 0.125            & 0.883               & 0.789                      & 0.56                    & 0.75                     \\ \hline
\textbf{MPT}    & Zero-shot & 0.146       & 0.072       & 0.044       & 0.027       & 0.092           & 0.123            & 0.886               & 0.824                      & 0.521                   & 0.71                     \\ \hline
\textbf{}       & One-shot  & 0.242       & 0.191       & 0.162       & 0.143       & 0.162           & 0.242            & 0.893               & 0.772                      & 0.503                   & 0.714                    \\ \hline
\textbf{}       & Two-shot  & 0.221       & 0.155       & 0.123       & 0.111       & 0.146           & 0.252            & 0.892               & 0.773                      & 0.498                   & 0.721                    \\ \hline
\end{tabular}
\end{adjustbox}
\caption{Results for generic automatic evaluation of \textbf{MultiWOZ} Dataset}
\label{gen_auto_woz}
\end{table*}

\subsection{Results for task oriented evaluation and analysis}\label{ResultsAndErrorAnalysis}

\begin{table}[ht]
\centering
\begin{adjustbox}{max width=\linewidth}
\begin{tabular}{|c|c|c|c|c|c|c|c|}
\hline
\textbf{Model}                   & \textbf{\begin{tabular}[c]{@{}c@{}}Few-shot\\ Setting\end{tabular}} & \textbf{Fluency} & \textbf{Adequacy} & \textbf{\begin{tabular}[c]{@{}c@{}}Contextual\\ Consistency\end{tabular}} & \textbf{Con} & \textbf{Pol} & \textbf{Emp} \\ \hline
\multirow{3}{*}{\textbf{Llama}}  & Zero-shot                                                           & 2.83             & 2.91              & 2.74                                                                      & 2.88         & 3.11         & 2.64         \\ \cline{2-8} 
                                 & One-shot                                                            & 3.53             & 3.47              & 3.32                                                                      & 3.42         & 3.61         & 3.37         \\ \cline{2-8} 
                                 & Two-shot                                                            & 3.93             & 3.63              & 3.52                                                                      & 3.79         & 3.81         & 3.96         \\ \hline
\multirow{3}{*}{\textbf{OPT}}    & Zero-shot                                                           & 3.42             & 3.20              & 3.13                                                                      & 3.12         & 3.21         & 3.10         \\ \cline{2-8} 
                                 & One-shot                                                            & 3.61             & 3.32              & 3.32                                                                      & 3.76         & 3.79         & 3.18         \\ \cline{2-8} 
                                 & Two-shot                                                            & 2.57             & 2.73              & 2.64                                                                      & 2.98         & 2.59         & 2.19         \\ \hline
\multirow{3}{*}{\textbf{Falcon}} & Zero-shot                                                           & 1.13             & 1.23              & 1.91                                                                      & 1.15         & 1.27         & 1.38         \\ \cline{2-8} 
                                 & One-shot                                                            & 1.58             & 2.01              & 1.74                                                                      & 1.76         & 1.74         & 1.89         \\ \cline{2-8} 
                                 & Two-shot                                                            & 1.35             & 1.78              & 1.59                                                                      & 1.42         & 1.66         & 1.75         \\ \hline
\multirow{3}{*}{\textbf{Alpaca}} & Zero-shot                                                           & 4.04             & 3.75              & 3.95                                                                      & 4.07         & 4.11         & 4.17         \\ \cline{2-8} 
                                 & One-shot                                                            & 3.63             & 3.29              & 3.36                                                                      & 3.43         & 3.68         & 3.83         \\ \cline{2-8} 
                                 & Two-shot                                                            & 3.21             & 2.98              & 3.14                                                                      & 3.25         & 3.34         & 3.46         \\ \hline
\multirow{3}{*}{\textbf{MPT}}    & Zero-shot                                                           & 4.32             & 3.96              & 4.15                                                                      & 4.36         & 4.53         & 4.23         \\ \cline{2-8} 
                                 & One-shot                                                            & 4.11             & 3.74              & 3.95                                                                      & 4.13         & 4.25         & 4.12         \\ \cline{2-8} 
                                 & Two-shot                                                            & 3.41             & 3.14              & 3.22                                                                      & 3.63         & 3.81         & 3.67         \\ \hline
\end{tabular}

\end{adjustbox}
\caption{Results of human evaluation on \textsc{MHLCD} dataset.}
\label{human_results_mhlcd}
\end{table}

\textbf{MHLCD} \ref{human_results_mhlcd}\ref{gen_auto_mhlcd}: LLaMA outperforms the others with the best performance, while Falcon performs the worst. LLaMA excels in two-shot settings, while OPT achieves the highest scores in one-shot settings. Its performance decreases in two-shot settings. Falcon, on the other hand, performs slightly better in one-shot settings compared to zero-shot and two-shot settings for all automatic evaluation metrics. OPT achieves the best scores in a one-shot setting. In two-shot settings, OPT's performance degrades.

\begin{table}[ht]
\centering
\begin{adjustbox}{max width=\linewidth}
\begin{tabular}{|l|l|l|l|l|l|l|l|}
\hline
\textbf{}       &           & NegStr & \textbf{BarStr} & \textbf{Fluency} & \textbf{\begin{tabular}[c]{@{}l@{}}Context\\ Relevance\end{tabular}} & \textbf{\begin{tabular}[c]{@{}l@{}}Negotiation\\ consistency\end{tabular}} & \textbf{\begin{tabular}[c]{@{}l@{}}Dialogue\\ Engagingness\end{tabular}} \\ \hline
\textbf{Llama2} & Zero-shot & 0.72   & 0.47            & 4.2              & 3.35                                                                 & 0.65                                                                       & 0.95                                                                     \\ \hline
\textbf{}       & One-shot  & 0.72   & 0.47            & 4.2              & 3.18                                                                 & 0.61                                                                       & 0.95                                                                     \\ \hline
\textbf{}       & Two-shot  & 0.71   & 0.44            & 4.2              & 3.44                                                                 & 0.64                                                                       & 0.94                                                                     \\ \hline
\textbf{OPT} & Zero-shot & 0.54   & 0.22            & 4.3              & 3.14                                                                 & 0.44                                                                       & 0.81                                                                     \\ \hline
\textbf{}       & One-shot  & 0.55   & 0.33            & 4.3              & 3.18                                                                 & 0.45                                                                       & 0.91                                                                     \\ \hline
\textbf{}       & Two-shot  & 0.51   & 0.33            & 4.3              & 3.13                                                                 & 0.44                                                                       & 0.9                                                                      \\ \hline
\textbf{Falcon}    & Zero-shot & 0.67   & 0.58            & 4.2              & 3.52                                                                 & 0.64                                                                       & 0.95                                                                     \\ \hline
\textbf{}       & One-shot  & 0.67   & 0.58            & 4.2              & 3.41                                                                 & 0.64                                                                       & 0.95                                                                     \\ \hline
\textbf{}       & Two-shot  & 0.64   & 0.54            & 4.2              & 3.46                                                                 & 0.62                                                                       & 0.95                                                                     \\ \hline
\textbf{ALPACA} & Zero-shot & 0.66   & 0.39            & 4.2              & 3.22                                                                 & 0.49                                                                       & 0.88                                                                     \\ \hline
\textbf{}       & One-shot  & 0.61   & 0.35            & 4.2              & 3.09                                                                 & 0.42                                                                       & 0.88                                                                     \\ \hline
\textbf{}       & Two-shot  & 0.62   & 0.35            & 4.2              & 3.41                                                                 & 0.57                                                                       & 0.89                                                                     \\ \hline
\textbf{MPT}    & Zero-shot & 0.68   & 0.41            & 4.2              & 3.12                                                                 & 0.61                                                                       & 0.93                                                                     \\ \hline
\textbf{}       & One-shot  & 0.61   & 0.38            & 4.2              & 3.41                                                                 & 0.62                                                                       & 0.92                                                                     \\ \hline
\textbf{}       & Two-shot  & 0.62   & 0.37            & 4.2              & 3.39                                                                 & 0.62                                                                       & 0.94                                                                     \\ \hline
\end{tabular}
\end{adjustbox}
\caption{Results of task-specific evaluation and human evaluation on \textsc{Craigslist Bargain} dataset.}
\label{results_cb}
\end{table}

\textbf{Craigslist Bargain\ref{results_cb}\ref{gen_auto_cb}}: For this dataset, the evaluation primarily focused on negotiation capability and some degree of logical understanding of the product when suggesting the price. The Falcon model performed the best in this case, with the Falcon one-shot model outperforming other variants in all automatic metrics except for NegStr, where Llama performs better. For our combined testing focusing primarily on task-specific metrics over the generic ones, the best model is Falcon, which performed the best, closely followed by the Llama model, with MPT, Alpaca, and OPT following in order. OPT was the model least likely to suggest a price in negotiation. All models accept Alpaca gained performance by few shot samples on average, with one shot giving the biggest boost. For the Negotiation task, Alpaca shows little change in response structure, leading to comparatively low improvements with few-shot samples, and only adjusts recommended seller prices. Alpaca's response structure remains consistent with few shots, except for recommended pricing, while with OPT, responses were often generated without price.

\begin{table}[ht]
\centering
\begin{adjustbox}{max width=\linewidth}
\begin{tabular}{|c|c|c|c|c|c|c|c|c|}
\hline
\textbf{Model}                   & \textbf{\begin{tabular}[c]{@{}c@{}}Few-shot\\ Setting\end{tabular}} & \textbf{Fluency} & \textbf{\begin{tabular}[c]{@{}c@{}}Contextual\\ Consistency\end{tabular}} & \textbf{Per} & \textbf{Emp} & \textbf{DonPr} & \textbf{PerStr} & \textbf{EmoPr}\\ \hline
\multirow{3}{*}{\textbf{Llama}}  & Zero-shot                                                           & 3.68                           & 3.55                                                                      & 3.04         & 2.68         & 0.58  & 52.45\%  & 38.66\%        \\ \cline{2-9} 
                                 & One-shot                                                            & 3.77                           & 3.61                                                                      & 3.56         & 3.01         & 0.68 
                                 & 58.28\% & 39.81\% 
                                 \\ \cline{2-9} 
                                 & Two-shot                                                            & 3.98                           & 3.84                                                                      & 3.69         & 3.23         & 0.71  
                                 & 60.27\%  & 42.64\% \\ \hline
\multirow{3}{*}{\textbf{OPT}}    & Zero-shot                                                           & 3.08                           & 2.86                                                                      & 3.56         & 3.47         & 0.62     & 59.98\% & 52.31\%           \\ \cline{2-9} 
                                 & One-shot                                                            & 3.11                           & 2.88                                                                      & 3.89         & 3.55         & 0.67 
                                 & 62.88\%  & 58.69\%
                                 \\ \cline{2-9} 
                                 & Two-shot                                                            & 3.16                           & 2.92                                                                      & 3.86         & \textbf{3.72}         & 0.66  
                                 & 62.08\%  & \textbf{66.21\%}
                                 \\ \hline
\multirow{3}{*}{\textbf{Falcon}} & Zero-shot                                                           & 2.98                           & 2.76                                                                      & 3.02         & 3.21         & 0.61     & 51.81\%                              & 44.86\%    \\ \cline{2-9} 
                                 & One-shot                                                            & 3.34                           & 3.02                                                                      & 3.43         & 3.36         & 0.67  
                                 & 56.69\%         & 49.87\% 
                                 \\ \cline{2-9} 
                                 & Two-shot                                                            & 3.16                           & 3.06                                                                      & 3.22         & 3.09         & 0.70    
                                 & 53.76\%       & 41.80\%
                                 \\ \hline
\multirow{3}{*}{\textbf{Alpaca}} & Zero-shot                                                           & 3.04                           & 2.66                                                                       & 3.52         & 2.87         & 0.62    & 59.16\%        & 40.66\%      \\ \cline{2-9} 
                                 & One-shot                                                            & \textbf{4.16}                           & \textbf{4.07}                                                                      & 3.87         & 2.98         & 0.72  
                                 & 62.42\%                              & 42.76\% \\ \cline{2-9} 
                                 & Two-shot                                                            & 3.88                           & 3.74                                                                      & \textbf{4.31}         & 3.12         & \textbf{0.78}  
                                 & \textbf{70.27\%}                             & 43.55\% \\ \hline
\multirow{3}{*}{\textbf{MPT}}    & Zero-shot                                                           & 3.16                           & 2.25                                                                      & 3.42         & 3.09         & 0.48    & 56.35\%                             & 41.56\%      \\ \cline{2-9} 
                                 & One-shot                                                            & 3.21                           & 2.46                                                                      & 3.12         & 3.35         & 0.50  & 52.35\%                             & 49.06\%       \\ \cline{2-9} 
                                 & Two-shot                                                            & 3.46                           & 2.30                                                                      & 3.36         & 3.21         & 0.52  
                                 & 54.61\%                              & 44.78\%
                                 \\ \hline
\end{tabular}

\end{adjustbox}
\caption{Results of task-specific and human evaluation on \textsc{Empathetic Persuasions} dataset.}
\label{human_results_p4g}
\end{table}

\textbf{Empathetic Persuasions\ref{human_results_p4g}\ref{gen_auto_ep}}: Overall, the LLMs performed best on the one-shot and two-shot prompts, followed by the zero-shot prompts. This is likely because the one-shot and two-shot prompts provide the LLM with more information about the task, which helps them to generate more accurate and persuasive responses.  
Speaking in the context of automatic generic evaluation, the LLMs performed better on the metrics that measure fluency and coherence (such as BLEU and METEOR) than on the metrics that measure semantic similarity (such as BertScore). This suggests that the LLMs are better at generating text that is fluent and coherent than at generating text that is semantically similar to the target text. The LLMs performed better on the metrics that measure the overall quality of the text (such as Embedding Average and Vector Extrema) than on the metrics that measure specific aspects of the text (such as Greedy Matching). This suggests that the LLMs are better at generating text that is of high overall quality than at generating text that meets specific criteria. The performance of the LLMs improved with the number of shots. This suggests that the LLMs benefit from more information about the task.
In the context of human evaluation metrics, Alpaca performed better the rest of the LLMs. Most of the models performed quite great in terms of fluency but were unable to retain the contextual consistency similarly. The donation probability increased as the model included persuasive and empathetic emotions in its responses. Again, the LLMs performed best on the one-shot and two-shot prompts, followed by the zero-shot prompts, similar to the results observed in automatic generic evaluation. PerStr and EmoPr were mostly in accordance with Per, Emp, and DonPr combined.

\begin{table}[ht]
\centering
\begin{adjustbox}{max width=\linewidth}
\begin{tabular}{|c|c|c|c|c|c|}
\hline
\textbf{Model}                   & \textbf{\begin{tabular}[c]{@{}c@{}}Few-shot\\ Setting\end{tabular}} & \textbf{Fluency} & \textbf{\begin{tabular}[c]{@{}c@{}}Contextual\\ Relevance\end{tabular}} & \textbf{Empathy} & \textbf{Emp}  \\ \hline
\multirow{3}{*}{\textbf{Llama}}  & Zero-shot                                                           & 3.72             & 3.27              & 3.09                                                                      & 51.32\%           \\ \cline{2-6} 
                                 & One-shot                                                            & 3.54             & 3.36              & 3.27                                                                     & 54.45\%       \\ \cline{2-6} 
                                 & Two-shot                                                            & 3.54            & 3.47              & 3.36                                                                     & 56.75\%              \\ \hline
\multirow{3}{*}{\textbf{OPT}}    & Zero-shot                                                           & 3.18            & 2.45             & 2.63                                                                     & 52.61\%       \\ \cline{2-6} 
                                 & One-shot                                                            & 3.81             & 3.63              & 3.02                                                                      & 57.28\%      \\ \cline{2-6} 
                                 & Two-shot                                                            & 3.72            & 3.54              & 3.45                                                                      & 64.67\%           \\ \hline
\multirow{3}{*}{\textbf{Falcon}} & Zero-shot                                                           & 3.61             & 3.29              & 3.45                                                                      & 62.16\%      \\ \cline{2-6}  
                                 & One-shot                                                            & 3.14             & 3.45              & 3.36                                                                      & 64.33\%       \\ \cline{2-6} 
                                 & Two-shot                                                            & 4.09            & 4.18              & 4.11                                                                     & 66.52\%             \\ \hline
\multirow{3}{*}{\textbf{Alpaca}} & Zero-shot                                                           & 3.27            & 3.45              & 3.09                                                                      & 50.58\%     \\ \cline{2-6}  
                                 & One-shot                                                            & 3.36             & 2.92              & 2.90                                                                     & 51.62\%       \\ \cline{2-6} 
                                 & Two-shot                                                            & 3.27             & 3.29             & 3.09                                                                      &53.56\%           \\ \hline
\multirow{3}{*}{\textbf{MPT}}    & Zero-shot                                                           & 3.63             & 3.64              & 3.27                                                                     & 54.44\%        \\ \cline{2-6}  
                                 & One-shot                                                            & 3.21             & 3.14              & 3.09                                                                      & 51.56\%       \\ \cline{2-6} 
                                 & Two-shot                                                            & 3.90             & 3.81              & 4.12                                                                      & 61.12\%         \\ \hline
\end{tabular}

\end{adjustbox}
\caption{Results of task-specific and human evaluation on \textsc{Empathetic Dialogues} dataset.}
\label{human_results_empathetic_dialogues}
\end{table}

\begin{table}[ht]
\centering
\begin{adjustbox}{max width=\linewidth}
\begin{tabular}{|c|l|l|l|l|l|l|l|l|}
\hline
\multicolumn{1}{|l|}{\textbf{}}  &           & \begin{tabular}[c]{@{}l@{}}Context\\  Consistency\\ (Greedy)\end{tabular} & \textbf{\begin{tabular}[c]{@{}l@{}}Context\\  Consistency\\ (Average)\end{tabular}} & \textbf{\begin{tabular}[c]{@{}l@{}}Context\\  Consistency\\ (Extrema)\end{tabular}} & \textbf{INFORM} & \textbf{SUCCESS} & \textbf{Fluency} & \textbf{\begin{tabular}[c]{@{}l@{}}Context \\ Relevance\end{tabular}} \\ \hline

\multirow{3}{*}{\textbf{Llama2}} & zero shot & 0.75                                                                   & 0.78                                                                             & 0.54                                                                             & 0.32            & 1.05             & 4.12             & 3.45                       \\ \cline{2-9} 
                                 & one shot  & 0.84                                                                   & 0.92                                                                             & 0.68                                                                             & 0.54            & 1.01             & 4.26             & 3.52                       \\ \cline{2-9} 
                                 & two shot  & 0.89                                                                   & 0.95                                                                             & 0.72                                                                             & 0.61            & 0.92             & 4.25             & 3.7                        \\ \hline
\multirow{3}{*}{\textbf{OPT}}    & zero shot & 0.47                                                                   & 0.53                                                                             & 0.4                                                                              & 0.25            & 1.44             & 4.1              & 2.97                       \\ \cline{2-9} 
                                 & one shot  & 0.54                                                                   & 0.67                                                                             & 0.49                                                                             & 0.38            & 1.38             & 4.16             & 3.22                       \\ \cline{2-9} 
                                 & two shot  & 0.6                                                                    & 0.71                                                                             & 0.55                                                                             & 0.43            & 1.08             & 4.17             & 3.28                       \\ \hline
\multirow{3}{*}{\textbf{Falcon}} & zero shot & 0.43                                                                   & 0.52                                                                             & 0.39                                                                             & 0.23            & 1.32             & 3.96             & 3.03                       \\ \cline{2-9} 
                                 & one shot  & 0.58                                                                   & 0.65                                                                             & 0.45                                                                             & 0.32            & 1.27             & 4.02             & 3.17                       \\ \cline{2-9} 
                                 & two shot  & 0.61                                                                   & 0.69                                                                             & 0.47                                                                             & 0.39            & 0.98             & 4.05             & 3.21                       \\ \hline
\multirow{3}{*}{\textbf{Alpaca}} & zero shot & 0.43                                                                   & 0.58                                                                             & 0.52                                                                             & 0.29            & 1.67             & 3.86             & 3.12                       \\ \cline{2-9} 
                                 & one shot  & 0.48                                                                   & 0.66                                                                             & 0.58                                                                             & 0.36            & 1.48             & 3.93             & 3.26                       \\ \cline{2-9} 
                                 & two shot  & 0.49                                                                   & 0.64                                                                             & 0.6                                                                              & 0.45            & 1.13             & 3.95             & 3.33                       \\ \hline
\multirow{3}{*}{\textbf{MPT}}    & zero shot & 0.69                                                                   & 0.72                                                                             & 0.51                                                                             & 0.33            & 1.25             & 4.2              & 3.39                       \\ \cline{2-9} 
                                 & one shot  & 0.8                                                                    & 0.88                                                                             & 0.61                                                                             & 0.41            & 1.18             & 4.21             & 3.53                       \\ \cline{2-9} 
                                 & two shot  & 0.83                                                                   & 0.91                                                                             & 0.69                                                                             & 0.56            & 0.93             & 4.18             & 3.74                       \\ \hline
\end{tabular}
\end{adjustbox}
\caption{Results of task-specific and human evaluation on \textsc{MultiWOZ} dataset.}
\label{human_results_woz}
\end{table}

\textbf{Empathetic Dialogues\ref{human_results_empathetic_dialogues}\ref{gen_auto_ed}}: We can see that across different scenarios, Falcon consistently emerges as the top performer. It demonstrates strong fluency, relevance, and empathy, with the highest scores in most categories. Falcon shows significant improvement as more shots are given, particularly in empathy, and has the highest percentage of users perceiving empathy in its responses.

MPT also stands out as a reliable performer, especially in terms of relevance and empathy. It shows substantial improvement from zero-shot to one-shot and two-shot scenarios, indicating its ability to adapt and generate more accurate and empathetic responses with additional training data. OPT shows promise with its adaptability and improvement from zero-shot to one-shot and two-shot scenarios, particularly in terms of fluency and relevance. Llama and Alpaca perform decently but demonstrate slightly lower scores compared to Falcon and MPT.

Overall, Falcon exhibits the strongest performance across fluency, relevance, and empathy, followed closely by MPT. These models showcase the potential for generating high-quality responses in different scenarios, with Falcon leading the way in terms of overall performance and user-perceived empathy.

\textbf{MultiWOZ\ref{human_results_woz}\ref{gen_auto_woz}}: Here, we can observe from the Table (Generic Automatic Evaluation) that all the Language Models have a similar kind of performance on the MultiWOZ dataset regarding Generic Metrics. But when it comes to Task-Specific metrics for Context Consistency, Llama2 and MPT perform better than the rest of the Language Models, and Llama2 and MPT are quite comparable in terms of Context Consistency. The performance of Llama2 and MPT differs for the INFORM metric. Llama2 performs way better than the rest of the Language Models when we evaluate the responses of the Model on a turn-level instead of a dialogue level and evaluate how considerate the model is towards the information provided by the user. Similarly, for the SUCCESS metric, Llama2 is a clear winner. This shows that Llama2 is better than the rest in providing the requested information by the user.

\subsection{Good Examples}\label{GoodExamples}

\subsubsection{Empathetic Persuasions}

\begin{itemize}
    \item  
    \textbf{Alpaca Zero-shot :}\\
    \textbf{Context :}\\ 
        \textit{user :} So the charity helped the Syrian children? \\
        \textit{bot :} Oh yes and others all over the globe. Would you donate to this charity today? \\
        \textit{user :} Yeah, maybe so. Can i donate part of the \$0.30 from this HIT or do I have to go to their website?\\

        \textbf{Gold response :} You can donate the \$0.30 form this hit every penny helps.\\
        \textbf{Generated response :} Yes, you can donate part of the \$0.30 from this HIT.

    \item
    \textbf{LLaMA2 One-shot :}\\
    \textbf{Context :}\\ 
        \textit{bot :} Great!  If you want, you can donate some of your task payment today.  Would you be interested in doing that?\\
\textit{user :} No thank you. I donate to other organizations on a yearly basis already. \\
\textit{bot :} That's great.  I certainly understand.  Save the Children does a lot to help children internationally.  Any amount you could give would help.  But I understand if you don't want to\\
\textit{user :} Thank you for understanding. I wish I were in a position to be able to donate more annually, but that simply isn't the case.\\
\textit{bot :} Yeah, I get that.  Asking someone to donate from their mTurk earnings is asking a lot, considering how little these HITs pay!\\

\textit{bot :} No judgment is ever passed. Save the Children is an international non-governmental organization that promotes children's rights, provides relief and helps support children in developing countries. URL Visit for more info.\\
\textit{user :} Thank you very much. I will look into that for sure. Perhaps in the future I will be able to make a donation.\\
\textit{bot :} I just want to be sure you do not wish me to ask How much do you like to donate to the charity now? Your donation will be directly deducted from your task payment. You can choose any amount from \$0 to all your payment (\$2) Please understand I must do my part as well.\\
\textit{user :} At this point I'm not able to donate anything so \$0.\\

        \textbf{Gold response :} I hope great things come for you and your loved ones. Thanks for your consideration..\\
        \textbf{Generated response :} I understand.  Would you like to donate to another charity?
\end{itemize}

\subsection{Bad Examples}\label{BadExamples}

\subsubsection{Empathetic Persuasions}

\begin{itemize}
    \item 
    \textbf{Falcon Zero-shot :}\\
    \textbf{Context :}\\ 
        \textit{bot :} There is a very important charity called Save the Children.  Do you know much about it? \\
        \textit{user :} I do not, can you tell me some? \\
        \textit{bot :} They help children all over who are in need.  They have programs to help with help, literacy, and disaster relief. \\
        \textit{user :} I would like to donate \\

        \textbf{Gold response :} How much of the money you earn from this task would you be willing to donate?\\
        \textbf{Generated response :} Thank you for your donation.

        \item
    \textbf{OPT One-shot :}\\
    \textbf{Context :}\\ 
        \textit{bot :} It's called Save the Children. Have you heard of it?\\
\textit{user :} I think so.  Does it have commercials on tv? \\
\textit{bot :} Yes, they do. I wish they didn't have to spend their monies on commercials but it is the only way to get their name out there.\\
\textit{user :} Does this charity help children in the United States, or do they deal mostly with children in African nations?\\
\textit{bot :} It is a global charity. They help out children in the US and around the world. You should check out their website:  URL s/what-we-do/us-programs\\

\textit{bot :} There is a very important charity called Save the Children.  Do you know much about it?\\
\textit{user :} I do not, can you tell me some?\\
\textit{bot :} They help children all over who are in need.  They have programs to help with help, literacy, and disaster relief. \\
\textit{user :} I would like to donate\\

        \textbf{Gold response :} How much of the money you earn from this task would you be willing to donate?\\
        \textbf{Generated response :} Hey you wanna donate some money today?
\end{itemize}

\begin{table*}[ht]
\centering
\begin{adjustbox}{max width=\linewidth}
\begin{tabular}{|l|p{2cm}p{2cm}p{15cm}|}
\hline
\textbf{Model}                  & \multicolumn{3}{c|}{\textbf{Examples}}                                                                                                                                                                                                                                                                                                                                                \\ \hline
\multirow{12}{*}{\textbf{LLaMA}} & \multicolumn{2}{c|}{\multirow{2}{*}{\textit{\textbf{Context}}}}                                                                                                                 & victim: Yes, My mother would feel relaxed and motivated to take action against mishappening happened to her, if she get any sort of counseling assistance.                                          \\ \cline{2-4} 
                                & \multicolumn{2}{c|}{\multirow{3}{*}{\textit{\textbf{Gold Response}}}}                                                                                                           & counselor: Yeah sure, we can help you with this. Do you want to connect with the National Commission for Women or Women-in-Distress helpline no. for getting counseling assistance for your mother? \\ \cline{2-4} 
                                & \multicolumn{1}{c|}{\multirow{8}{*}{\textit{\textbf{\begin{tabular}[c]{@{}c@{}}Predicted\\ Response\end{tabular}}}}} & \multicolumn{1}{l|}{\textbf{Zero-shot}} & We would love to help you with the counseling options for your mother. Would you like to seek counseling?                                                                                           \\ \cline{3-4} 
                                & \multicolumn{1}{c|}{}                                                                                                & \multicolumn{1}{l|}{\textbf{One-shot}}  & counselor:  It's very nice to hear that you are very concerned about your mother's health.  We have counseling options for your mother to help her recover from the trauma of being scammed.  Would you like to seek counseling for your mother? \\ \cline{3-4} 
                                & \multicolumn{1}{c|}{}                                                                                                & \multicolumn{1}{l|}{\textbf{Two-shot}}  & {counselor:  We are very glad to hear that. I would like to inform you that we have a team of professional counselors and social workers who are trained to provide counseling to victims of online abuse. They can help you and your mother to deal with the trauma and the stress that you have experienced.} \\ \hline
\multirow{12}{*}{\textbf{OPT}} & \multicolumn{2}{c|}{\textit{\textbf{Context}}}                                                                                                                 & victim: What information do I have to submit here for filing the complaint? \\ \cline{2-4} 
                               & \multicolumn{2}{c|}{\textit{\textbf{Gold Response}}}                                                                                                           & counselor: You have to give your email and contact number as a information. \\ \cline{2-4} 
                               & \multicolumn{1}{c|}{\multirow{8}{*}{\textit{\textbf{\begin{tabular}[c]{@{}c@{}}Predicted\\ Response\end{tabular}}}}} & \multicolumn{1}{l|}{\textbf{Zero-shot}} & counselor: National cyber crime portal - The portal has two steps: Step 1. Report a cybercrime Step 2. Get Help The first step would entail you to upload a scan copy of the communication you have with the cyber criminal from where ever you can find. \\ \cline{3-4} 
                               & \multicolumn{1}{c|}{}                                                                                                & \multicolumn{1}{l|}{\textbf{One-shot}}  & {counselor: The first step is to type your complaint in the box provided. Enter your complaint number followed by the description of the crime here. In case you do not know your complaint number, you can submit details about the incident to us and our system will retrieve the information for you.} \\ \cline{3-4} 
                               & \multicolumn{1}{c|}{}                                                                                                & \multicolumn{1}{l|}{\textbf{Two-shot}}  & counselor: Report a cybercrime and upload a scan copy of the communication you have with the cyber criminal from where ever you can find. \\ \hline
\multirow{5}{*}{\textbf{Falcon}} & \multicolumn{2}{c|}{\textit{\textbf{Context}}}                                                                                                                 & victim: I need help to file a complaint against the company that scammed my mother. \\ \cline{2-4} 
                               & \multicolumn{2}{c|}{\textit{\textbf{Gold Response}}}                                                                                                           & counselor: I’m sorry to hear that your mother has been scammed. To file a complaint, you will need to provide some personal information, such as your email and contact number, and details about the incident. \\ \cline{2-4} 
                               & \multicolumn{1}{c|}{\multirow{3}{*}{\textit{\textbf{\begin{tabular}[c]{@{}c@{}}Predicted\\ Response\end{tabular}}}}} & \multicolumn{1}{l|}{\textbf{Zero-shot}} & counselor: I can guide you through the process of filing a complaint. First, you’ll need to gather any evidence you have, like messages or emails. \\ \cline{3-4} 
                               & \multicolumn{1}{c|}{}                                                                                                & \multicolumn{1}{l|}{\textbf{One-shot}}  & counselor: It’s important to report this scam to the authorities. You can start by providing your mother’s details and a brief description of what happened. I can help you with the next steps. \\ \cline{3-4} 
                               & \multicolumn{1}{c|}{}                                                                                                & \multicolumn{1}{l|}{\textbf{Two-shot}}  & counselor: Thank you for reaching out. To help your mother, please gather all the relevant information and documents. We’ll walk you through the complaint process. \\ \hline
\multirow{5}{*}{\textbf{Alpaca}} & \multicolumn{2}{c|}{\textit{\textbf{Context}}}                                                                                                                 & victim: My mother was scammed online, and I need help. What should I do? \\ \cline{2-4} 
                               & \multicolumn{2}{c|}{\textit{\textbf{Gold Response}}}                                                                                                           & counselor: I'm really sorry to hear that. It's important to act quickly. You can start by reporting the scam to the police and collecting all the evidence you have. \\ \cline{2-4} 
                               & \multicolumn{1}{c|}{\multirow{3}{*}{\textit{\textbf{\begin{tabular}[c]{@{}c@{}}Predicted\\ Response\end{tabular}}}}} & \multicolumn{1}{l|}{\textbf{Zero-shot}} & counselor: The first step is to report the scam. You can provide the details to the authorities and make sure you keep all the evidence safe. \\ \cline{3-4} 
                               & \multicolumn{1}{c|}{}                                                                                                & \multicolumn{1}{l|}{\textbf{One-shot}}  & counselor: It’s good that you’re taking action. You should file a report with the local authorities and contact your bank if any financial details were involved. \\ \cline{3-4} 
                               & \multicolumn{1}{c|}{}                                                                                                & \multicolumn{1}{l|}{\textbf{Two-shot}}  & counselor: I understand how distressing this must be. Please report the incident to the police and provide any information that can help with the investigation. \\ \hline
\multirow{5}{*}{\textbf{MPT}} & \multicolumn{2}{c|}{\textit{\textbf{Context}}}                                                                                                                 & victim: My mother has been scammed, and I don’t know what to do. \\ \cline{2-4} 
                               & \multicolumn{2}{c|}{\textit{\textbf{Gold Response}}}                                                                                                           & counselor: I'm sorry to hear that. You should start by reporting the scam to the police and your bank. Also, keep all the evidence, such as emails and messages. \\ \cline{2-4} 
                               & \multicolumn{1}{c|}{\multirow{3}{*}{\textit{\textbf{\begin{tabular}[c]{@{}c@{}}Predicted\\ Response\end{tabular}}}}} & \multicolumn{1}{l|}{\textbf{Zero-shot}} & counselor: First, gather all the information you have about the scam. Then, you should report it to the authorities. \\ \cline{3-4} 
                               & \multicolumn{1}{c|}{}                                                                                                & \multicolumn{1}{l|}{\textbf{One-shot}}  & counselor: It’s important to act quickly. Please report the scam to the relevant authorities and your bank. Make sure you have all the details ready. \\ \cline{3-4} 
                               & \multicolumn{1}{c|}{}                                                                                                & \multicolumn{1}{l|}{\textbf{Two-shot}}  & counselor: I’m here to help. Start by reporting the scam to the police. Keep all the evidence you have, such as emails or texts. \\ \hline

\end{tabular}
\end{adjustbox}
\caption{Sample cases of good responses generated by the different LLMs for \textsc{mhlcd} dataset.}
\label{good_ex_mhlcd}
\end{table*}

\begin{table*}[ht]
\centering
\begin{adjustbox}{max width=\linewidth}
\begin{tabular}{|l|p{2cm}p{2cm}p{15cm}|}
\hline
\textbf{Model}                  & \multicolumn{3}{c|}{\textbf{Examples}}                                                                                                                                                                                                                                                                                                                                                \\ \hline
\multirow{5}{*}{\textbf{LLaMa}} & \multicolumn{2}{c|}{\textit{\textbf{Context}}}                                                                                                                 & victim: No I have no idea. \\ \cline{2-4} 
                               & \multicolumn{2}{c|}{\textit{\textbf{Gold Response}}}                                                                                                           & counselor: Would you please share if you have faced this issue before? \\ \cline{2-4} 
                               & \multicolumn{1}{c|}{\multirow{3}{*}{\textit{\textbf{\begin{tabular}[c]{@{}c@{}}Predicted\\ Response\end{tabular}}}}} & \multicolumn{1}{l|}{\textbf{Zero-shot}} & counselor: Do you have any idea who... \\ \cline{3-4} 
                               & \multicolumn{1}{c|}{}                                                                                                & \multicolumn{1}{l|}{\textbf{One-shot}}  & counselor: Can you share... \\ \cline{3-4} 
                               & \multicolumn{1}{c|}{}                                                                                                & \multicolumn{1}{l|}{\textbf{Two-shot}}  & counselor: Would you please be comfortable sharing for how long you have been...? \\ \hline
\multirow{5}{*}{\textbf{OPT}} & \multicolumn{2}{c|}{\textit{\textbf{Context}}}                                                                                                                 & victim: He is threatening me to be in a relationship with him for the rest of my life while he is married. \\ \cline{2-4} 
                               & \multicolumn{2}{c|}{\textit{\textbf{Gold Response}}}                                                                                                           & counselor: Would you please share why he is doing this to you if he is married? \\ \cline{2-4} 
                               & \multicolumn{1}{c|}{\multirow{3}{*}{\textit{\textbf{\begin{tabular}[c]{@{}c@{}}Predicted\\ Response\end{tabular}}}}} & \multicolumn{1}{l|}{\textbf{Zero-shot}} & counselor: I am worried about who he might hurt next. \\ \cline{3-4} 
                               & \multicolumn{1}{c|}{}                                                                                                & \multicolumn{1}{l|}{\textbf{One-shot}}  & counselor: I am worried about who he might hurt next; if you do not trust and tell me the truth, I will have no alternative but to report to the police. The sooner, the better.  \\ \cline{3-4} 
                               & \multicolumn{1}{c|}{}                                                                                                & \multicolumn{1}{l|}{\textbf{Two-shot}}  & counselor: I see. \\ \hline
\multirow{5}{*}{\textbf{Falcon}} & \multicolumn{2}{c|}{\textit{\textbf{Context}}}                                                                                                                 & victim: What kinds of details does she have to provide while getting this assistance here? \\ \cline{2-4} 
                               & \multicolumn{2}{c|}{\textit{\textbf{Gold Response}}}                                                                                                           & counselor: You will have to provide your Indian registered mobile number and email ID. \\ \cline{2-4} 
                               & \multicolumn{1}{c|}{\multirow{3}{*}{\textit{\textbf{\begin{tabular}[c]{@{}c@{}}Predicted\\ Response\end{tabular}}}}} & \multicolumn{1}{l|}{\textbf{Zero-shot}} & counselor: Your name, sister. \\ \cline{3-4} 
                               & \multicolumn{1}{c|}{}                                                                                                & \multicolumn{1}{l|}{\textbf{One-shot}}  & counselor: Your sister, to for and hosting their your... \\ \cline{3-4} 
                               & \multicolumn{1}{c|}{}                                                                                                & \multicolumn{1}{l|}{\textbf{Two-shot}}  & counselor: Being your... \\ \hline
\multirow{5}{*}{\textbf{Alpaca}} & \multicolumn{2}{c|}{\textit{\textbf{Context}}}                                                                                                                 & victim: I want to file a complaint against someone who scammed my mother. \\ \cline{2-4} 
                               & \multicolumn{2}{c|}{\textit{\textbf{Gold Response}}}                                                                                                           & counselor: I understand. Could you please provide some details about the scam? \\ \cline{2-4} 
                               & \multicolumn{1}{c|}{\multirow{3}{*}{\textit{\textbf{\begin{tabular}[c]{@{}c@{}}Predicted\\ Response\end{tabular}}}}} & \multicolumn{1}{l|}{\textbf{Zero-shot}} & counselor: It's important to file a report with the police. \\ \cline{3-4} 
                               & \multicolumn{1}{c|}{}                                                                                                & \multicolumn{1}{l|}{\textbf{One-shot}}  & counselor: Please file a complaint as soon as possible. \\ \cline{3-4} 
                               & \multicolumn{1}{c|}{}                                                                                                & \multicolumn{1}{l|}{\textbf{Two-shot}}  & counselor: File a report with the police immediately. \\ \hline
\multirow{5}{*}{\textbf{MPT}} & \multicolumn{2}{c|}{\textit{\textbf{Context}}}                                                                                                                 & victim: I need help reporting a scam. My mother has been defrauded. \\ \cline{2-4} 
                               & \multicolumn{2}{c|}{\textit{\textbf{Gold Response}}}                                                                                                           & counselor: I'm sorry to hear that. You should report this to the authorities and your bank right away. \\ \cline{2-4} 
                               & \multicolumn{1}{c|}{\multirow{3}{*}{\textit{\textbf{\begin{tabular}[c]{@{}c@{}}Predicted\\ Response\end{tabular}}}}} & \multicolumn{1}{l|}{\textbf{Zero-shot}} & counselor: Make sure to keep all the evidence. \\ \cline{3-4} 
                               & \multicolumn{1}{c|}{}                                                                                                & \multicolumn{1}{l|}{\textbf{One-shot}}  & counselor: Report the scam to the authorities. \\ \cline{3-4} 
                               & \multicolumn{1}{c|}{}                                                                                                & \multicolumn{1}{l|}{\textbf{Two-shot}}  & counselor: Please contact the police immediately. \\ \hline
\end{tabular}
\end{adjustbox}
\caption{Sample cases of bad responses generated by the different LLMs for \textsc{mhlcd} dataset.}
\label{bad_ex_mhlcd}
\end{table*}

\section{Conclusion}\label{Conclusion}

In this work, we evaluated 5 prevalent LLMs Llama \cite{touvron2023Llama}, OPT \cite{zhang2022opt}, Falcon \cite{zxhang2023falcon}, Alpaca \cite{alpaca} and MPT \cite{MosaicML2023Introducing} across a diverse range of tasks, namely reservation, empathetic response generation, mental health and legal counseling, persuasion, and negotiation. We selected datasets corresponding to the tasks in focus to evaluate the capabilities of LLMs in a conversational setting. Through a rigorous evaluation process, we conclude that no LLM is optimal for all the tasks, and the choice of LLM depends on the task. Some models were better for one task but performed relatively poorly in other specific tasks.

In conclusion, our study provides a comprehensive evaluation of five prevalent LLMs across a diverse range of conversational tasks. We found that no LLM is overall optimal for all tasks, highlighting the importance of considering task-specific requirements when selecting an LLM. Additionally, we identified drawbacks in the performance of LLMs, including the inconsistency across tasks and the reliance on the validity of gold responses. To ensure optimal functional performance, it is crucial to evaluate LMs based on the specific capabilities they demonstrate for each distinct task. By understanding their strengths and limitations within different conversational domains, researchers and practitioners can make informed decisions when choosing the most appropriate LM for specific applications.

\section*{CRediT authorship contribution statement}
\noindent \textbf{Ratnesh Kumar Joshi}: Writing – original draft, Conceptualization, Methodology, Software.  \textbf{Priyanshu Priya:} Software, Writing – original draft \& review, Investigation, Validation. \textbf{Vishesh Desai}: Software, Investigation, Validation. \textbf{Saurav Dudhate}: Software, investigation, Validation. \textbf{Siddhant Senapati}: Software, Investigation, Validation. \textbf{Asif Ekbal}: Methodology, Supervision, Writing – review. \textbf{Roshni Ramnani}: Writing – review, Investigation , Supervision. \textbf{Anutosh Maitra}: Writing – review, Conceptualization, Supervision. \textbf{Shubhashis Sengupta}:  Writing – review, Supervision.

\section*{Declaration of generative AI and AI-assisted technologies in the writing process}
\noindent During the preparation of this work, the author(s) used [ChatGPT:] to [paraphrase and summarize the text] and [Grammarly:] to [improve the manuscript's grammar, spelling, and sentence structure]. After using this tool/service, the author(s) reviewed and edited the content as needed and take(s) full responsibility for the publication's content.

\section*{Declaration of Competing Interest}
\noindent The authors declare that they have no known competing financial interests or personal relationships that could have appeared to influence the work reported in this paper. The authors declare the following financial interests/personal relationships that may be considered potential competing interests:  Ratnesh Kumar Joshi, Priyanshu Priya, and Asif Ekbal report a relationship with the Indian Institute of Technology Patna that includes: employment. Roshni Ramnani, and Anutosh Maitra reports a relationship with Technology Innovation Lab, Accenture, that provides for employment.

\section*{Acknowledgement}

\noindent The research reported in this paper is an outcome of the project titled \enquote{Conversational Agents with Negotiation and Influencing Ability}, sponsored by Accenture Labs, Banglore, India.

\bibliography{template}

\end{document}